\documentclass[nohyperref]{article}

\usepackage{microtype}
\usepackage{graphicx}
\usepackage{subfigure}
\usepackage{booktabs} 

\usepackage{multirow}

\usepackage{hyperref}

\usepackage[accepted]{icml2022}

\usepackage{amsmath}
\usepackage{amssymb}
\usepackage{mathtools}
\usepackage{amsthm}

\usepackage[capitalize,noabbrev]{cleveref}

\theoremstyle{plain}

\theoremstyle{definition}

\theoremstyle{remark}

\usepackage[textsize=tiny]{todonotes}

\icmltitlerunning{Expression might be enough: representing pressure and demand for reinforcement learning based traffic signal control}

\begin{document}

\twocolumn[\icmltitle{Expression might be enough: representing pressure and demand for reinforcement learning based traffic signal control}



\icmlsetsymbol{equal}{*}

\begin{icmlauthorlist}
\icmlauthor{Liang Zhang}{lzu}
\icmlauthor{Qiang Wu}{uestc}
\icmlauthor{Jun Shen}{wol}
\icmlauthor{Linyuan Lü}{uestc}
\icmlauthor{Bo Du}{wol2}
\icmlauthor{Jianqing Wu}{jiangxi}
\end{icmlauthorlist}

\icmlaffiliation{lzu}{School of Life Sciences, Lanzhou University, Lanzhou 730000, China}
\icmlaffiliation{uestc}{Institute of Fundamental and
    Frontier Sciences, University of Electronic Science and Technology of China, Chengdu 611731, China}
\icmlaffiliation{wol}{School of Computing and Information Technology, University of Wollongong, Wollongong, Australia}
\icmlaffiliation{wol2}{SMART Infrastructure Facility, University of Wollongong, Wollongong, Australia}
\icmlaffiliation{jiangxi}{School of Information Engineering, Jiangxi University of Science and Technology}

\icmlcorrespondingauthor{Qiang Wu}{qiang.wu@uestc.edu.cn}

\icmlkeywords{Reinforcement learning, Represent learning, Application}

\vskip 0.3in
]



\printAffiliationsAndNotice{\icmlEqualContribution} 

\begin{abstract}
Many studies confirmed that a proper traffic state representation is more important than complex algorithms for the classical traffic signal control (TSC) problem.
%
In this paper, we (1) present a novel, flexible and efficient method, namely advanced max pressure (Advanced-MP), taking both running and queuing vehicles into consideration to decide whether to change current signal phase; (2) inventively design the traffic movement representation with the efficient pressure and effective running vehicles from Advanced-MP, namely advanced traffic state (ATS); and (3) develop a reinforcement learning (RL) based algorithm template, called Advanced-XLight\footnote{https://github.com/LiangZhang1996/Advancd\_XLight}, by combining ATS with the latest RL approaches, and generate two RL algorithms, namely "Advanced-MPLight" and "Advanced-CoLight" 
from Advanced-XLight.
Comprehensive experiments on multiple real-world datasets show that:  (1) the Advanced-MP outperforms baseline methods, and it is also efficient and reliable for deployment; and (2) Advanced-MPLight and Advanced-CoLight can achieve the state-of-the-art.
%

\end{abstract}

\section{Introduction}
Traffic signal control (TSC) is essential for improving transportation efficiency and mitigating traffic congestion.
In many cities, the classical FixedTime~\cite{fixedtime}, SCOOT~\cite{scoot2}, and SCATS~\cite{scats2} are still the most commonly deployed TSC systems.
Max pressure (MP) control~\cite{mp2013} and self-organizing traffic lights (SOTL)~\cite{sotl2013} aim to maximize the global throughput from observation of traffic states.

%
These conventional methods cannot be easily adapted to complex dynamic traffic flows without experts' prior knowledge. Recently, reinforcement learning (RL) has drawn increasing interests. 
Intuitively, RL-based TSC methods ~\cite{drl,frap} can directly learn from the complex conditions through trail-and-reward, without requiring specific, which are often unrealistic, assumptions about the traffic model. 
Some RL-based methods have shown superior performance over many traditional methods in certain situations, which can also be used to control large-scale traffic signals ~\cite{presslight,mplight,colight}.

%

%
Assume the observation of the number of a lane is $n$.
When all signal phases (detailed in Section 3) are  considered, the complexity of exploration  state space for RL can raise to $O(n^{8})$~\cite{frap}.
Hence, the complex RL-based approaches might not be the optimal solutions for TSC. On the other hand, both traffic state and neural network design play essential roles in RL-based methods for TSC. 
%
%
The success of both MPLight~\cite{mplight} and Efficient-CoLight~\cite{efficient} has shown that combining and optimizing traffic state representations from traditional methods with RL-based models could yield significant improvements, especially in large-scale TSC.

The traffic movement pressure expressed by efficient pressure (EP)~\cite{efficient} is an essential representation of traffic state. 
Although EP has strong representativeness, it ignores the running vehicles in the traffic network. Here, the "pressure" alone is not sufficient to represent the complex traffic state. Meanwhile, with a greedy strategy,  Efficient-MP~\cite{efficient} is not flexible.

To summarize, the limitations of current traffic state representation and TSC methods are listed as follows:
\begin{itemize}
    \item Traditional TSC methods with a greedy strategy are still not flexible (e.g., they cannot maintain current phases when there are multiple running vehicles, while other competing phases have very few queuing vehicles);
    \item Most current traffic state representations used in the RL-based approaches neglect the running vehicles in the traffic network;
    \item State-of-the-art RL-based TSC methods still have room for improvement with more effective traffic state representations.
\end{itemize}
To further address the challenges, we have considered combining the running vehicles and queuing vehicles for traffic state design optimization and leveraged the "request" concept from the actuated control method SOTL~\cite{sotl2013}. 
Hence, the main contributions of this paper are as follows:
\begin{enumerate}
    \item We present a novel, flexible and efficient method, called advanced max pressure (Advanced-MP), taking both running  and queuing vehicles into consideration  to decide whether to change the current signal phase;
    \item We design the advanced traffic state (ATS), which combines the pressure and effective running vehicles as the traffic state representation;
    \item We develop an RL-based algorithm template, namely Advanced-XLight, which combines ATS with any RL approaches and generates two RL algorithms: Advanced-MPLight and Advanced-CoLight, for instance;
    \item We demonstrate that our Advanced-MP method and the two RL algorithms generated could achieve new state-of-the-art (SOTA) for large-scale TSC.
\end{enumerate}

\section{Related Work}
Various TSC approaches have been proposed, which can be divided into two typical categories: traditional approaches and RL-based methods. 
\subsection{Traditional approaches}
Back in 1958, FixedTime method~\cite{fixedtime} specified a fixed cycle length and phase split for each traffic light phase.
Subsequently, SCOOT~\cite{scoot2}, SCATS~\cite{scats2}, MP control~\cite{mp2013}, and SOTL~\cite{sotl2013}, among others, have tried to set the signal according to the road traffic conditions, which are still widely used in most cities.

The concept of MP control ~\cite{origin1990} was initially developed for scheduling packets in wireless communication networks.
Varaiya et al.~\cite{mp2013} and  Gregoire et al.~\cite{bp20142} formulated the definition of max-pressure and proved its stability. 
Adapted for signalized intersections, some studies on MP~\cite{mp2012, bp2014, bp20142, mp2014} have demonstrated its efficiency through simulation.  
Le et.al~\cite{mp2015} and  Levin et.al~\cite{mp2020} proposed cyclical phase structure, which had greater palatability for implementation.
The Efficient-MP method~\cite{efficient} proposed a simple but efficient traffic state representation (one "lanes to lanes" approach to calculate pressure) based on MP, and achieved SOTA, even among the latest RL-based approaches. 

Despite the high performance of the MP-based control, it lacks flexibility and needs an optimal hyper-parameters set, such as action duration and cycle length. 
For the implementation of MP applications for TSC, some works set a fixed duration for each activated phase~\cite{mp2012, bp2014, bp20142, presslight, mplight, survey}, while some studies used fixed cycle length and then proportionally set phase split~\cite{mp2014, mp2015, mp2020}. 

With a fixed action duration or cycle length, the MP control cannot maintain a phase when a platoon of vehicles is passing through the intersection, because the "pressure" only concentrates on the queuing vehicles and ignores the running vehicles.
For example, considering there are no running vehicles at the current phase, but many queuing vehicles being at other phases, the phase signal should be changed, and the MP control can make the right decision; while there are many running vehicles at current phase, but fewer queuing vehicles at other phases, the phase signal should not be changed, MP control cannot work properly because it only concentrates on the queuing vehicles. 

Self-organizing traffic lights control (SOTL) ~\cite{sotl2004,sotl2012-2, sotl2013} is one type of adaptive TSC method, which can autonomously adjust the phase signal according to the traffic state.
It evaluates the conditions of the green phase and other competing phases, and adaptively decides whether to maintain or change the current signal phase.
The main advantage of the SOTL is its flexibility to dynamic traffic flow. 
Besides, the SOTL pays attention to the running vehicles crossing through, which may also be essential for traffic optimization. 
We will leverage the advantages of SOTL into Efficient-MP~\cite{efficient} to make it more adaptive, in this paper.

\subsection{RL-based methods}

So far the RL-based methods have shown excellent performance, mainly for two reasons: 1) advanced deep neural networks; 2) well-designed state and reward strategy. 

FRAP~\cite{frap} proposed a novel network structure realized by phase competition and relation, which can deal with unbalanced traffic flow and have strong transferability. 
CoLight~\cite{colight} used graph attention networks (GAT)~\cite{gats} to realize intersection level cooperation and is capable of handling large scale traffic signal control.
PressLight~\cite{presslight} integrated "pressure" into state reward design, and accomplished multi-intersection TSC.
Obviously, integrating the concept of pressure into the RL-based methods could bring significant improvement. 
MPLight~\cite{mplight} used FRAP as the base neural network to control city-level traffic signals. 
Efficient-CoLight~\cite{efficient} introduced "efficient pressure" (EP) and uses its traffic state representational ability to achieve the SOTA. 
The representation of traffic state  will have a strong impact on the quality of the TSC models.

In this paper, our work aims to answer three questions: (1) how to improve Efficient-MP's performance and make it more adaptive to dynamic traffic? (2) how to represent the traffic state more effectively? and (3) how to further improve the performance of the RL-related methods without adding complexity?

\section{Preliminaries}
In this section, we summarize relevant definitions~\cite{mplight,efficient} for TSC in this paper.

\begin{figure}[htbp]
	\includegraphics[width=1\linewidth]{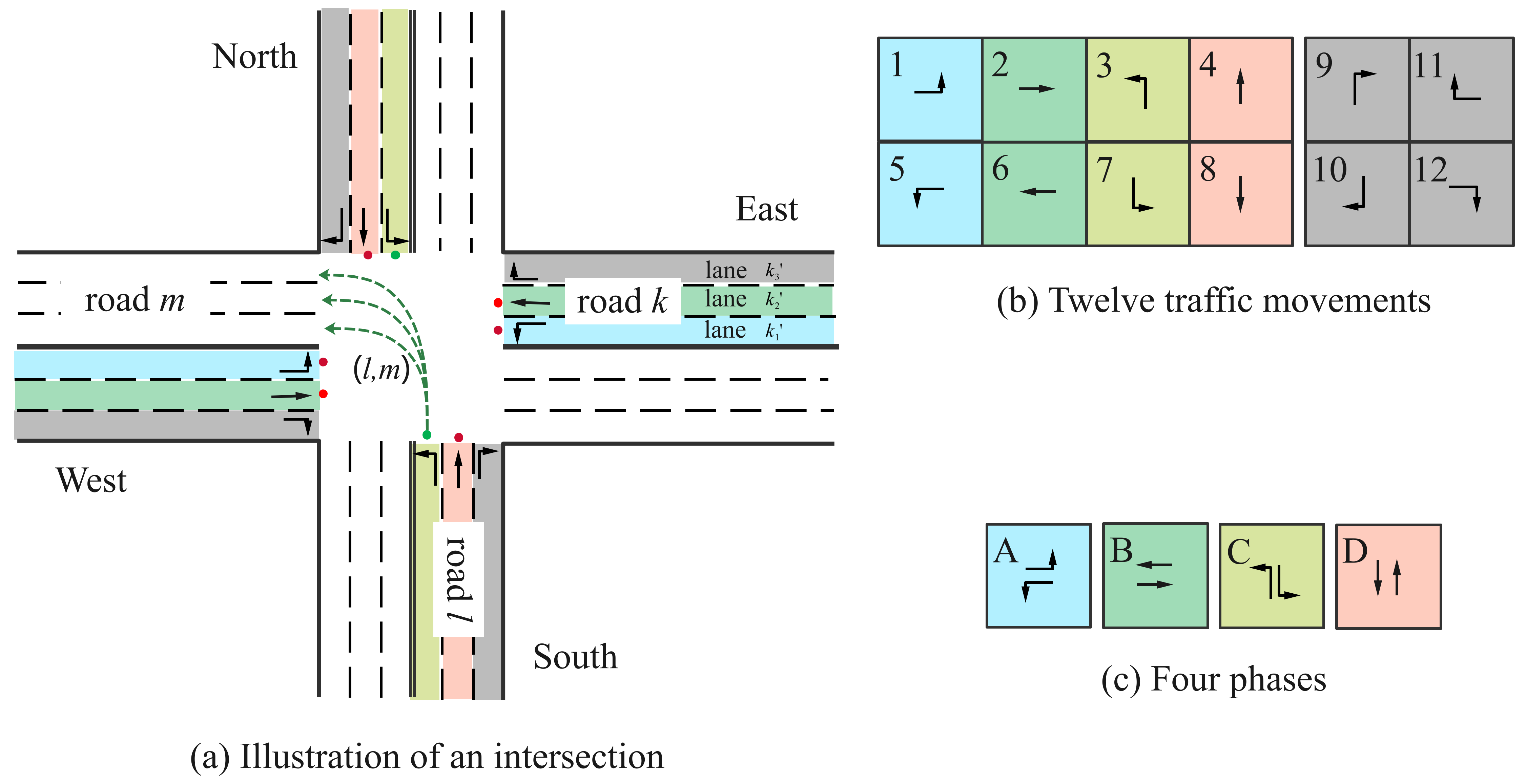}
	\caption{An illustration of intersection, traffic movements and traffic signal phase.}
	\label{fig:traffic}
\end{figure}

\textbf{Definition 1} (Traffic network). 
A traffic network is described as a directed graph, in which each node represents the intersection, and each edge represents the road. 
Each intersection has incoming roads and outgoing roads, and each road consists of several lanes which determine how the vehicle pass through the intersection, such as turn left, go straight, and turn right. 
An incoming lane for an intersection is where the vehicle enters the intersection. An outgoing lane for an intersection is where the vehicle leaves the intersection. We denote the set of incoming lanes and outgoing lanes of intersection $i$ as $\mathcal{L}_{i}^{in}$ and $\mathcal{L}_{i}^{out}$, respectively. We use $l,m,k$ to denote the roads, and  $l', m',k'$ to denote the lanes. Figure~\ref{fig:traffic}(a) illustrates an intersection and eight roads.

\textbf{Definition 2} (Traffic movement). Traffic movement is defined as the traffic traveling across an intersection from one incoming road to one outgoing road. We denote a traffic movement from road $l$ to road $m$ as $(l,m)$, in which $(l,m)=set\{(l', m')\},l'\in l, m'\in m$. For an intersection with each road having three lanes, for instance, each traffic movement contains one entering lane and three exiting lanes. Figure~\ref{fig:traffic}(a) uses three green dash lines to describe the traffic from the south to west, and there are a total of  twelve traffic movements.

\textbf{Definition 3} (Signal phase). Each signal phase is a set of permissible traffic movements (as shown in Figure  ~\ref{fig:traffic}(b)). We denote one phase with $s$, in which $s=set\{(l, m)\}$, and $s\in \mathcal{S}_i$. Figure  ~\ref{fig:traffic}(c) describes the mostly used four phases.

\textbf{Definition 4} (Efficient pressure). The efficient pressure of each traffic movement is the average difference of queue length between the upstream and downstream, denoted by 
\begin{equation}
	e(l,m)=  \frac{1}{M}\sum_{i=1}^{M} q(l'_i)- \frac{1}{N}\sum_{j=1}^{N}q(m'_j), l' \in l, m' \in m
\end{equation}
in which $q(l')$ represents the queue length of lane $l'$, and $M$ and $N$ are the used lane numbers of road $l$ and $m$, respectively.

\textbf{Definition 5} (Phase pressure). The pressure of each phase is the sum of efficient pressure of the traffic movements, which form the phase, denoted by 
\begin{equation}
	p(s) = \sum e(l, m), (l, m) \in s, s\in \mathcal{S}_i
\end{equation}
in which $e(l, m)$ represents the traffic movement's efficient pressure and it is calculated by equation (1).

\textbf{Definition 6} (Intersection pressure). The pressure of each intersection is defined as the difference of queue length between the upstream and downstream, denoted by 
\begin{equation}
P_i = \sum q(l')-\sum q(m'), l' \in \mathcal{L}_{i}^{in}, m' \in \mathcal{L}_{i}^{out}
\end{equation}
in which $q(l')$ represents the queue length of lane $l'$.

\textbf{Definition 7} (Action duration). The action duration of each agent is denoted as $t_{duration}$, and the phase duration can be $n*t_{duration}, n\in\mathbb{N}^+$.

The summary of notations is listed in Table~\ref{tab:notation}.
\begin{table}[htb]
	\caption{Summary of notations.}
	\label{tab:notation}
	\begin{tabular}{ll}
	\toprule
	Notation & Description\\
	\midrule
	$\mathcal{L}_{i}$  & set of lanes of intersection $i$\\
	$\mathcal{L}_{i}^{in}$ & set of incoming lanes of intersection $i$\\
	$\mathcal{L}_{i}^{out}$ & set of outgoing lanes of intersection $i$\\
	$l', m', k'$ & lanes  \\ 
	$l, m, k$ & road which is a set of lanes\\
	$q(l')$ & queue length of lane $l'$\\ 
	$(l,m)$ &  a traffic movement from road $l$ to $m$\\
	$e(l,m)$ & efficient pressure from road $l$ to $m$\\
	$s$ & phase which is a set of traffic movements\\
	$\mathcal{S}_i$ & phases of intersection $i$\\
	$p(s)$ & pressure of phase $s$\\
	$P_i$ & pressure of intersection $i$\\
	$t_{duration}$ & action duration of each agent\\
	\midrule
    $V_{max}$ & the maximum velocity of vehicles\\
	$L$ & the effective range  within $t_{duration}$\\
    $r_e(l')$ & the running vehicle number within $L$\\
	$d(s)$ & the demand of phase $s$\\
    $ATS(l,m)$ & advanced traffic state for  $(l,m)$\\
	\bottomrule
	\end{tabular}
\end{table}

\textbf{Problem 1} (Multi-intersection traffic signal control). Each intersection is controlled by one RL agent. At time step $t$, agent $i$ views the environment as its observation $o_t^i$. Every $t_{duration}$, the action $a_t^i$ is taken to control the signal of intersection $i$. The goal of the agent is to take an optimal action $a_t^i$ (i.e. which phase to set) to maximize the throughput of the systems and minimize the average travel time.

\section{Method}
In this section, we develop the Advanced-MP based on Efficient-MP and SOTL control.
Next, we design the novel traffic movement representation with pressure and demand from Advanced-MP, namely advanced traffic state (ATS).
Finally, we introduce ATS into RL-base models and develop an template called Advanced-XLight.

\subsection{Advanced Max Pressure Control}
To make Efficient-MP control more adaptive, inspired by SOTL, we propose advanced max pressure (Advanced-MP), which takes both pressure and running vehicles near the intersection into consideration.
Advanced-MP can be regarded as an approximation of optimal MP, which sets dynamic phase duration for each phase under dynamic traffic.

\subsubsection{Definitions}
To depict the demand precisely, we herein firstly define some new concepts:

\textbf{Definition 8} (Effective range). The effective range is the maximum distance to the intersection that a vehicle can pass through  within $t_{duration}$, denoted by 
\begin{equation}
    L = V_{max} \times t_{duration}
\end{equation}
in which $V_{max}$ is the maximum velocity of vehicles. For example, if the vehicles' maximum velocity is $11 m/s$, then $L=110m$ for $t_{duration}=10s$, and $L=165m$ for $t_{duration}=15s$. The effective range is similar to $\omega$ (the distance to the intersection) in SOTL but with a more precise and deterministic value. 

\textbf{Definition 9} (Effective running vehicle number). The effective running vehicle number is the number of running vehicles of the incoming lanes within the effective range to the intersection, denoted by
\begin{equation}
    r(l,m)=\sum r_e(l'), l' \in l
\end{equation}
 in which $r_e(l')$ is the running vehicle number within $L$  of lane $l'$.

\textbf{Definition 10} (Phase demand). The demand of each phase is the sum of the effective running vehicle number from the phase, denoted by 
\begin{equation}
d(s) = \sum r(l,m), (l,m)\in s  
\end{equation}
in which $r(l, m)$ is calculated by equation (6).

The phase demand precisely explains the demand for passing through the intersection. For the current phase, phase demand can represent the need to maintain the current phase for the next $t_{duration}$, while phase pressure can represent the need to change the current phase. When the current phase pressure is too small, there may be congestion on the downstream. 

Phase demand represents the need for running vehicles, and phase pressure expresses the need for queuing vehicles. They compete for a green signal. For the current phase, we use $W_1$ to represent the importance weight of phase demand, and the "request" from the current phase is denoted by:
\begin{equation}
    d(\hat{s})\times W_1 
\end{equation}
Moreover, the "request" from other competing phases is $max\left(p(s)\right)$. We need to compare the "request" values and activate the phase with the maximum value.

In general, we propose the phase demand (\textbf{Definition 10}) under effective range (\textbf{Definition 8}) to quantify the request from the current phase, while the phase's efficient pressure is used as the request from other competing phases.

\subsubsection{Algorithm}
With Advanced-MP, not all the running vehicles have the demand (or request) for green signals because only those running on the effective range can pass through the intersection for the next $t_{duration}$. 

\begin{algorithm}[htp]
	\caption{Advanced Max-Pressure control}
	\label{alg:ap}
	\textbf{Parameter}: time $t=0$, action duration $t_{duration}$, weights $W_1$, current phase $a_{cur}$
	\begin{algorithmic}
	\STATE For each intersection, get  $p(s)$
	    \STATE $a_{cur}=$ arg max $\left(p(s)|, s\in \mathcal{S}_i\right)$
		\FOR{(time-step)}
		\STATE $t=t+1$
		\IF{$t=$$t_{duration}$}
		\STATE For each intersection, get  $p(s)$ and  $d(s)$
		\IF{$d(a_{cur})\times W_1 > max\{p(s)\}$}
		\STATE Maintain the current phase 
		\ELSE 
		\STATE $a_{cur}=$ arg max $\left(p(s)|, s\in \mathcal{S}_i\right)$ 
		\ENDIF
		\STATE Set the phase as $a_{cur}$.
		\STATE $t=0$
		\ENDIF
		
		\ENDFOR
	\end{algorithmic}
\end{algorithm}

The advanced MP method is formally summarized in Algorithm~\ref{alg:ap}.
With a fixed action duration, we test several simple weights and finally find the best as the final results of this method. For the phases with red signal, phase pressure can represent the need to be set for the next $t_{duration}$, while phase demand does not work. 
For each phase, phase pressure and phase demand cannot work together to represent the need for the phase. The phase demand and pressure can represent the "request" of the current phase and other competing phases, respectively. Through the competition of phase pressure and phase demand, Advanced-MP can realize adaptive control.

\subsubsection{Discussion} 
One crucial research question is, whether Advanced-MP still stabilizes the queue length in the traffic. 
Herein we will compare Advanced-MP with those typical MP-based methods  using the proportional phase split. 
Firstly, Advanced-MP can be regarded as one approximate realization of MP using proportional phase split under optimal cycle length. Advanced-MP does not change the properties of MP but it is a more adaptive realization. 

It is supposed that MP-based methods could set an optimal phase duration with different traffic conditions, and this duration might vary with different pressures. To get the optimal duration, we need to check whether the system uses small time slots, and evaluate whether to change the phase or not according to the traffic state, and finally get the optimal duration through joining these slots. Obviously, that is what Advanced-MP has done.

From this perspective, Advanced-MP can also be considered as equivalent to setting an optimal phase duration for each activated phase. Clearly, it can stabilize the traffic flows like MP-based methods and achieve superior performance (see details in experiment section).

\subsection{Advanced Traffic State}
Although the efficient pressure~\cite{efficient} also has strong expression capability in representing the traffic states, it can not express the running vehicles in the traffic network. After all, "pressure" alone is not enough to represent the complex traffic state. There is always a  hypothesis \cite{TITS} that representation is critical for computational intelligence algorithms, however, in this paper we want to reconfirm whether 'expression might be enough'. 

Based on Advanced-MP, we further design the advanced traffic state (ATS) with efficient traffic movement pressure and effective running vehicles for traffic state representation.

\textbf{Definition 11} (Advanced traffic state). The combination of representation of the efficient pressure of vehicle queues and effective running vehicles for traffic movement $(l,m)$, named advanced traffic state (ATS), is denoted by:
\begin{equation}
ATS(l,m) =\{(e(l,m);r(l,m)\}
\end{equation}
is which $e(l,m)$ and $r(l,m)$ are computed by equation (1) and equation (6), respectively.

\subsection{Advanced-XLight} 
We now can develop an advanced RL-based methods template applying ATS as traffic state, namely Advanced-\textit{X}Light (as Algorithm \ref{Advanced-XLight} shows). 

MPLight~\cite{mplight} and CoLight~\cite{colight} are adopted as the Q-network baseline architecture of our method considering their high performance. We have generated Advanced-MPLight and Advanced-CoLight by our Advanced-\textit{X}Light algorithm template.
It should be noted that the idea of RL-based design is not limited to MPLight and CoLight, as it can also be integrated into other RL-based models.
\begin{algorithm}
	 \caption{Advanced-XLight}
	 \label{Advanced-XLight}
	 \textbf{Parameter}: Current phase time $t$, minimum action duration $t_{duration}$
	 \begin{algorithmic}
		 \FOR{(time-step)}
		 \STATE $t=t+1$;
		 \IF{$t=t_{duration}$} 
		 \STATE Get $ATS$ by equation (8) for each intersection;
		 \STATE Set the phase by \textit{X} RL model; 
		 \STATE $t=0$;
		 \ENDIF
		 \ENDFOR
	 \end{algorithmic}
 \end{algorithm}

 \paragraph{State representation} Each agent observes the current phase and advanced traffic state (\textbf{Definition 11}) which consists of traffic movement efficient pressure (\textbf{Definition 4}) and effective running vehicles (\textbf{Definition 9}).

 \paragraph{Action} At time $t$, each agent chooses a phase $\hat{s}$ as its action $a_t$, and the traffic signal will be set to phase $\hat{s}$.

\paragraph{Reward} For Advanced-MPLight model, the reward is the pressure of the intersection, denoted by $r_i=-|P_i|$. The agent tries to stabilize the queues in the system by maximizing the reward. For the Advanced-CoLight model, the reward is the total queue length of the intersection, $r_i = -\sum q(l'), l'\in \mathcal{L}_i^{in}$. The Advanced-CoLight agent tries to minimize the queue length of the system.

The Advanced-XLight is updated by the Bellman Equation:
\begin{equation}
	Q(s_t, a_t)  = R(s_r, a_t) + \gamma max Q(s_{t+1}, a_{t+1})
\end{equation}

\section{Experiment}
We conduct experiments on six real-world datasets to evaluate our proposed methods, especially the performance of Advanced-MP and the importance of ATS in TSC.

\begin{table*}[htb]
\caption{Performance (the average travel time in seconds) comparison of different methods evaluated on JiNan, HangZhou and New York real-world datasets (the smaller the better).}
\label{all-res}
\begin{center}
\begin{tabular}{lcccccc}
    \toprule
    \multirow{2}{*}{ Method } & \multicolumn{3}{c}{ JiNan } & \multicolumn{2}{c}{ HangZhou } & New York\\
    \cline{ 2 - 7 } & 1 & 2 & 3 & 1 & 2 & 1\\
    \midrule
    FixedTime & $428.11 $ & $368.77$ & $383.01$ & $495.57$ & $406.65$ & $1507.12$\\
    MaxPressure & $273.96$ & $245.38$ & $245.81$ & $288.54$ & $348.98$ & $1179.55$\\
    Efficient-MP& ${269.87}$ & ${239.75}$ & ${240.03}$ & ${284.44}$ & ${327.62}$ & ${1122.00}$\\
    \midrule 
    MPLight & $297.46$ & $270.05$ & $276.15$ & $314.60 $ & $357.61 $&$1321.40$ \\
    CoLight & $272.06 $ & $252.44 $ & $249.56 $ & $297.02 $ & $347.27 $ &$ 1065.64$\\
    AttendLight & $277.53 $ &$ 250.29$ &$248.82$ &$293.89 $ &$345.72 $ &$ 1586.09$ \\
    PRGLight & $ 291.27$ & $257.52 $&$261.74 $& $301.06 $&$369.98 $& $1283.37 $\\
    Efficient-MPLight & $261.81$ & $241.35$ & $238.80$ & $284.49$ & $\mathbf{321.08}$ & $1301.83$\\
    Efficient-CoLight & $\mathbf{2 5 6 . 8 4}$ & $\mathbf{2 3 9 . 5 8}$ & $\mathbf{2 3 6 . 7 2}$ & $\mathbf{2 8 2 . 0 7}$ & $324.27$ & $\mathbf{1032.11}$\\
    \midrule
    Advanced-MP & $\mathbf{253.61}$ & $\mathbf{238.62}$ & $\mathbf{235.21}$ & $\mathbf{279.47}$ & $\mathbf{318.67}$ & ${1060.41}$\\
    Advanced-MPLight & $\mathbf{251.29}$ & $\mathbf{234.78}$ & $\mathbf{231.76}$ & $\mathbf{273.26}$ & $\mathbf{312.68}$ & $1198.64$\\
    Advanced-CoLight & $\mathbf{2 4 5 . 7 3}$ & $\mathbf{2 3 2 . 6 3}$ & $\mathbf{2 2 9 . 0 1}$ & $\mathbf{2 7 0 . 4 5}$ & $\mathbf{3 1 0 . 7 4}$ & $\mathbf{970.05}$\\
    \bottomrule
\end{tabular}
\end{center}
\end{table*}

\subsection{Experiment Settings}
We conduct experiments on CityFlow~\cite{cityflow}\footnote{https://cityflow-project.github.io}, a simulator that supports large-scale traffic signal control. The simulator provides the state to the agent and receives the phase signal settings. Each green signal is followed by a three-second yellow signal and two-second all red time to prepare the transition.

The traffic dataset consists of the road network data and traffic flow data. The road network data describes the traffic road links, traffic movements, and corresponding signal settings.  In the traffic flow dataset, each vehicle is described as $(t, u)$ where $t$ is time, $u$ is the pre-planned route, and $u$ is a set of roads located on the road network from the origin location to the destination.

In multi-intersection TSC, the phase number and minimum action duration are important hyper-parameters and should be set as the same before conducting baseline. 
We set the phase number as four and minimum action duration as 15-second, the same as traffic signal settings from Efficient-MP~\cite{efficient}.
 
\subsection{Datasets}
We use six real-world traffic datasets\footnote{https://traffic-signal-control.github.io} in experiments. 

\textbf{JiNan datasets:}  The road network has 12 ($3\times4$) intersections. Each intersection is four-way, with two 400-meter(East-West) long road segments and two 800-meter(South-North)long road segments. There are three traffic flow datasets under this traffic road network dataset. 

\textbf{HangZhou datasets:} The road network  has 16 ($4\times4$) intersections. Each intersection is four-way, with two 800-meter(East-West) long road segments and two 600-meter(South-North)long road segments. There are two traffic flow datasets under this traffic road network dataset.

\textbf{New York dataset:} The road network has 192 ($28\times 7$) intersections. Each intersection is four-way, with two 300-meter(East-West) long road segments and two 300-meter (South-North) long road segments. There are only one traffic flow dataset under this traffic road network dataset.


These traffic datasets are not only different from the perspectives of traffic road network, but also the arrival patterns. 

\subsection{Evaluation Metrics}
Following the existing studies~\cite{survey}, we use the average travel time to evaluate the performance of different models for traffic signal control. It calculates the average travel time of all the vehicles spent between entering and leaving the traffic network (in seconds), which is the most frequently used measure of performance to control traffic signals~\cite{frap, colight}.

\subsection{Compared Methods}
We compare our methods with the following baseline methods, including traditional traffic and RL methods. All the RL methods are trained with the same hyper-parameters (learning rate, replay buffer size, sample size). Each episode is a 60-minute simulation, and we adopt results as the average of the last ten testing episodes. The final report result is the average of three independent results.

\textbf{Traditional Methods:}
\begin{itemize}
    \item \textbf{Fixed-Time}~\cite{fixedtime}: a policy gives a fixed cycle length with a predefined phase split among all the phases.
    \item \textbf{Max-Pressure}~\cite{mp2013}: the max pressure control selects the phase that has the maximum pressure. 
    \item \textbf{Efficient-MP}~\cite{efficient}: it selects the phase with the maximum efficient pressure. It is a SOTA method that has superior performance than  MPLight~\cite{mplight}.  
\end{itemize}

\textbf{RL Methods:}
\begin{itemize}
    \item \textbf{MPLight}~\cite{mplight}: using FRAP~\cite{frap} as the base model, and introduces pressure into the state and reward design. 
	\item \textbf{CoLight}~\cite{colight}: using graph attention network to realize intersection cooperation. 
    \item \textbf{AttendLight}~\cite{attendlight}: using attention mechanism~\cite{attention} to construct phase feature and predict phase transition probability.
    \item \textbf{PRGLight}~\cite{prglight}: using graph neural network to predict traffic state and adjusts the phase duration according to the currently observed traffic state and predicted state.
    \item \textbf{Efficient-MPLight}~\cite{efficient}: FRAP~\cite{frap} based model, using current phase and efficient traffic movement pressure as observation, intersection pressure as reward.
    \item \textbf{Efficient-CoLight}~\cite{efficient}: CoLight based model, using current phase and efficient traffic movement pressure as observation, intersection queue length as reward. It is the SOTA RL method. 
\end{itemize}


\subsection{Results}
\subsubsection{Weight setting}
\begin{figure}[h]
    \begin{center}
    \includegraphics[width=1\linewidth]{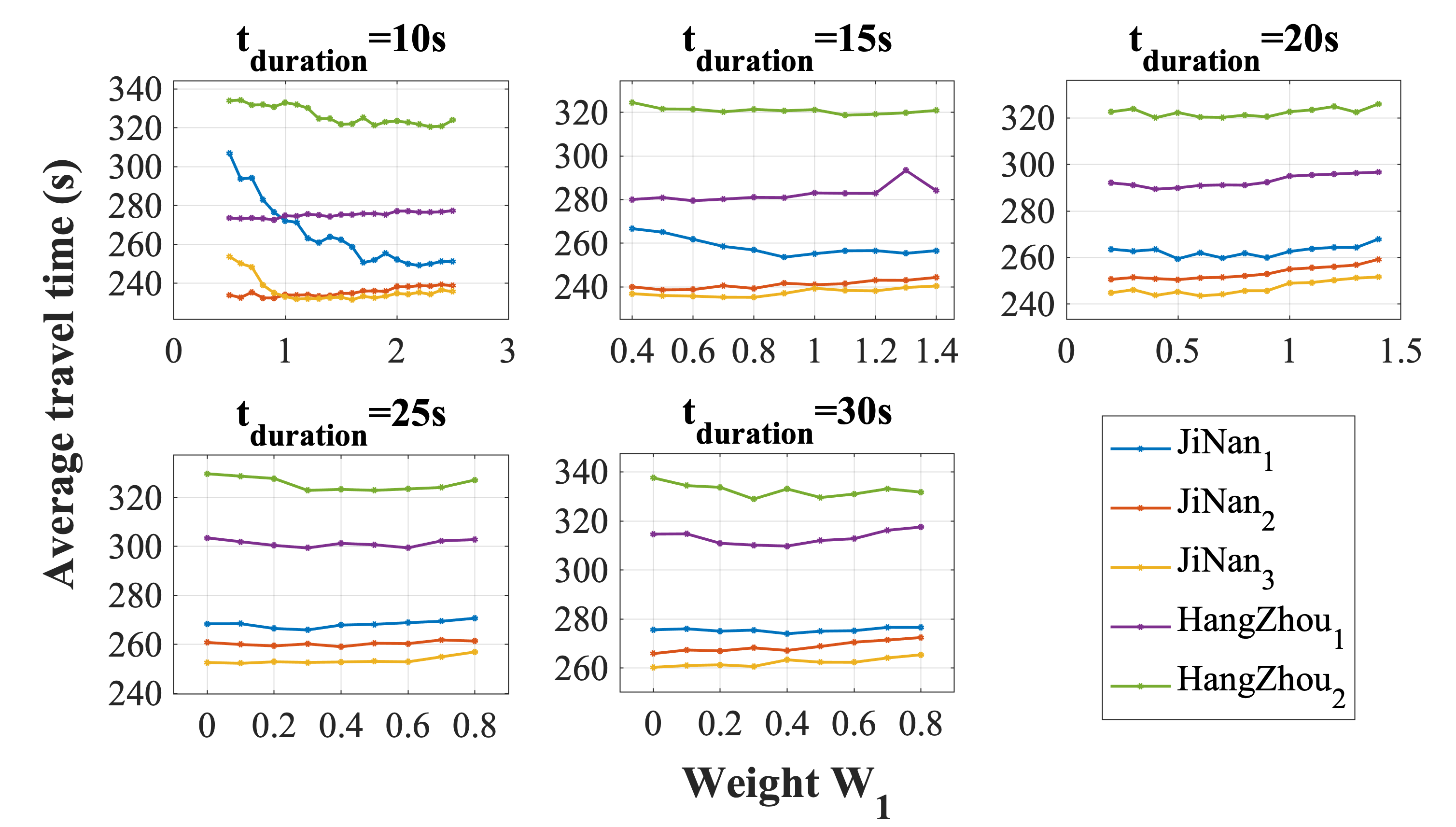}
    \caption{The performance of Advanced-MP under different weight $W_1$. }
    \label{fig:weight}
    \end{center}
\end{figure}

Various trials and testings are needed to get an appropriate $W_1$ for a particular action duration.
From our experiences (as shown in Figure~\ref{fig:weight}), the demand will need a higher weight with a smaller action duration, mainly for  two reasons:
(1) The time waste of phase transition (yellow time and all-red time) should be considered. The wasted time is fixed. Smaller action duration will potentially cause more frequent transition and more wasted time. 
(2) The smaller action duration will use a shorter efficient range (\textbf{Definition 8}), leading to a smaller value of traffic demand. Therefore the weight should set a higher value.
In other words, we should give a higher weight for demand with a small action duration and a lower one with a large action duration.

\subsubsection{Overall Performance}
Table~\ref{all-res} reports our experimental results under JiNan, HangZhou and New York
real-world datasets with respect to the average travel time. We have the following findings:

(1) Our proposed Advanced-MP consistently outperforms all other previous methods. 
With a well-designed weight $W_1$, Advanced-MP can outperform the SOTA RL method: Efficient-CoLight in JiNan and HangZhou datasets, and its performance is almost the same with Efficient-CoLight in the New York dataset.
Advanced-MP demonstrates the efficiency and adaptiveness of  transportation methods. 
Furthermore, compared to RL-based methods, Advanced-MP is easier to deploy because Advanced-MP only requires action duration and weight, and does not need intensive training.

(2) Our proposed Advanced-CoLight outperforms all other methods. Integrating AST with the RL-based approaches brings excellent improvements. The performance of our RL-based algorithm  (Advanced-CoLight) is improved by  $13.54\%$ and $6.01\%$ from the previous SOTA methods, Efficient-MP and Efficient-CoLight, respectively. 

\subsubsection{Variation with action duration}

Action duration (also considered as minimum phase duration, $t_{duration}$) is an essential hyper-parameter, and it influences the control performance of both transportation and RL-based methods.
To prove that our proposed methods have a better performance under different action duration, experiments under different action duration are also conducted.

\begin{figure}[htb]
    \centering
    \includegraphics[width=1\linewidth]{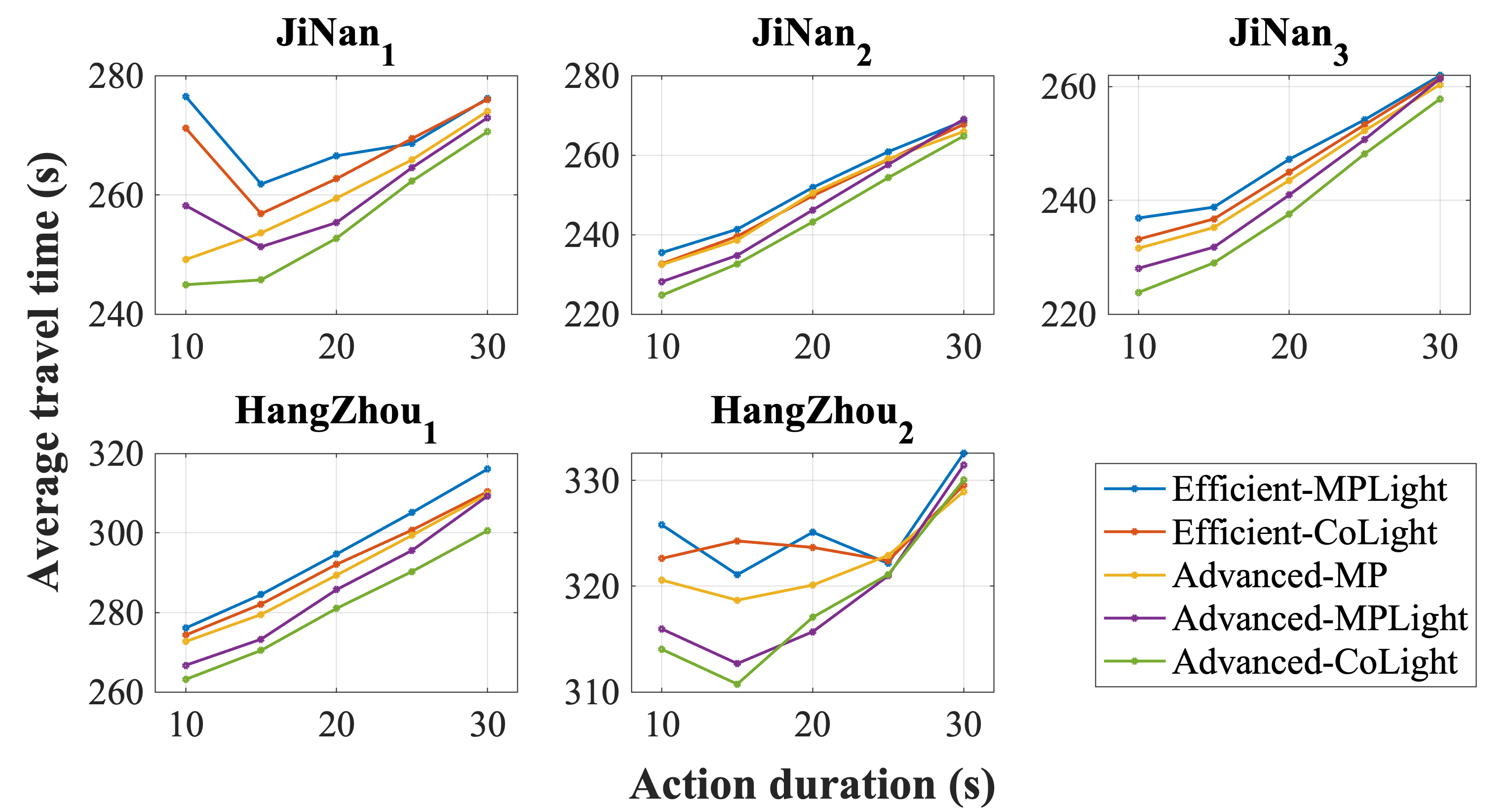}
    \caption{Model performance under different action duration, the smaller the better.}
    \label{fig:duration}
\end{figure}

Figure \ref{fig:duration} reports the model performance on JiNan and HangZhou datasets under different action duration. 
The performance of Advanced-MP gets a better control with a smaller action duration (except at $HangZhou_2$). This is in tune with the fact that it can be regarded as an approximation of optimal MP, which sets optimal phase duration all the time. The Advanced-MP can estimate optimal MP more accurately with a smaller phase duration because the controller can update traffic state more frequently and make more precise decisions.

We find that:
(1) Phase duration has a great influence on the model performance. All the methods have different model performances under different action durations. 
(2) Our proposed Advanced-MP outperforms the previous SOTA methods Efficient-CoLight over different action durations. Advanced-MP is more powerful than others.
(3) Our proposed Advanced-MPLight and Advanced-CoLight perform better than the previous SOTA methods. Integrating traffic demand truly brings efficient improvement.

\subsubsection{Effective observation}

Comprehensive experiments are conducted to address the importance of effective range and running vehicles.
Advanced-CoLight and Advanced-MPLight are evaluated under the three different state configurations with different observations. 
It differs from getting the number of running vehicles near the intersection: 
(1) In \textbf{Config1}, the current phase, traffic movement pressure, and the number of running vehicles in the incoming lanes within 100 meters near the intersection are used as the state representation;
(2) In \textbf{Config2}, the current phase, traffic movement pressure, and the number of running vehicles in the incoming lanes within 200 meters near the intersection are used as the state representation;
(3) In \textbf{Config3}, the current phase, traffic movement pressure, and  the number of all the running vehicles in each entering lane are used as the state representation;
(4) In \textbf{Default}, the current phase, traffic movement pressure, and the number of effective running vehicles are used as the state representation.
The model performances are compared with \textbf{Default} to evaluate the importance of effective range.

 \begin{figure}[H]
    \centering
     \includegraphics[width=1\linewidth]{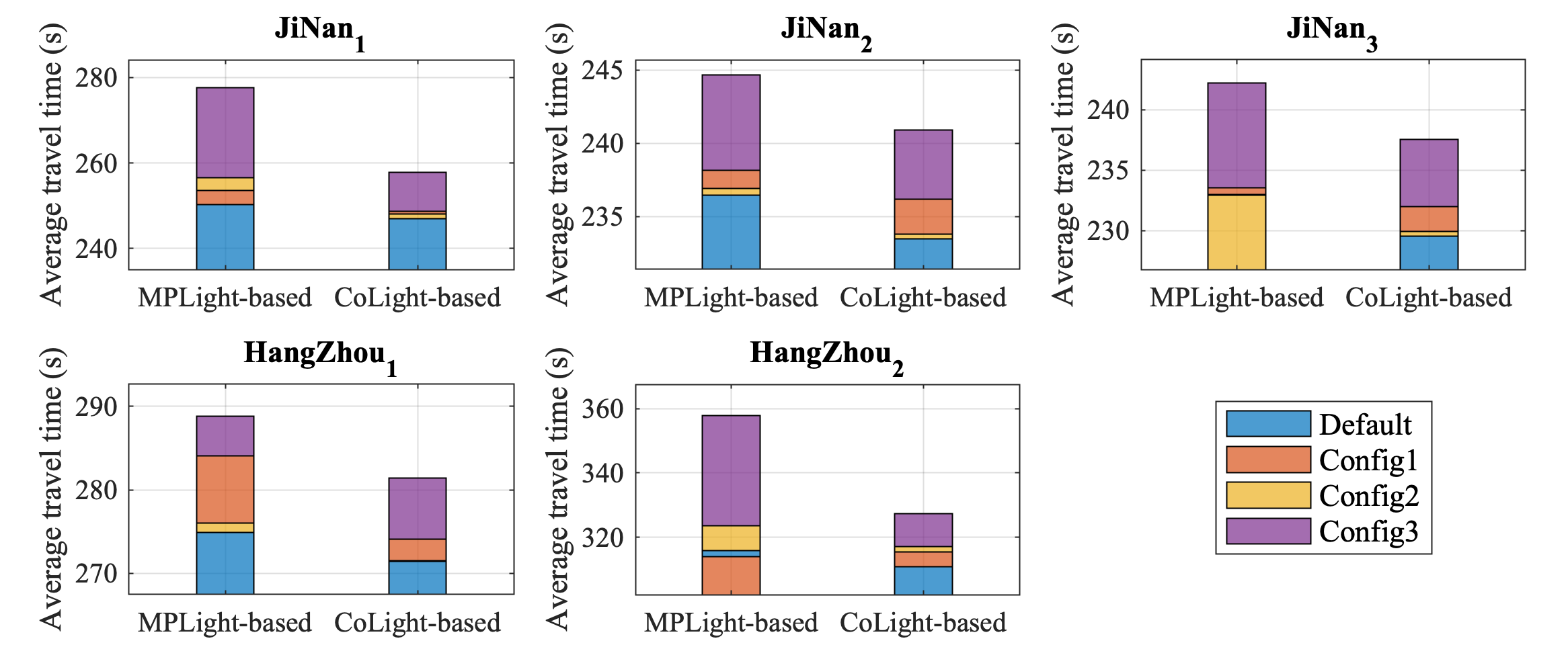}
     \caption{Model performance under different state representation}
     \label{fig:config}
\end{figure}

Figure \ref{fig:config} reports the model performance under different state configurations of observation range.
We find that  
(1) the observation range of running vehicles significantly influences the model performance. 
The performance is the best under the \textbf{Default} configuration.
(2) A model cannot perform better with more complex observations. 
Compared to Efficient-MPLight and Efficient-CoLight (Efficient-MPLight and Efficient-CoLight use current phase and traffic movement pressure as the state representation), the model gets worse performance under \textbf{Config3} with an additional observation of the running vehicles.

Therefore, the effective observation of running vehicles is essential for model improvement.
And the excellent performance of Advanced-CoLight and Advanced-MPLight should attribute to effective observation rather than more observations. 

\subsubsection{Model Generalization}

To evaluate the generalization of Advanced-XLight, we choose Advanced-MPLight while Advanced-CoLight has a limited capacity of transferability due to the influence of traffic network topologies.
We train Advanced-MPLight on JiNan and HangZhou datasets and transfer it to other datasets.

\begin{figure}[H]
    \centering
     \includegraphics[width=1\linewidth]{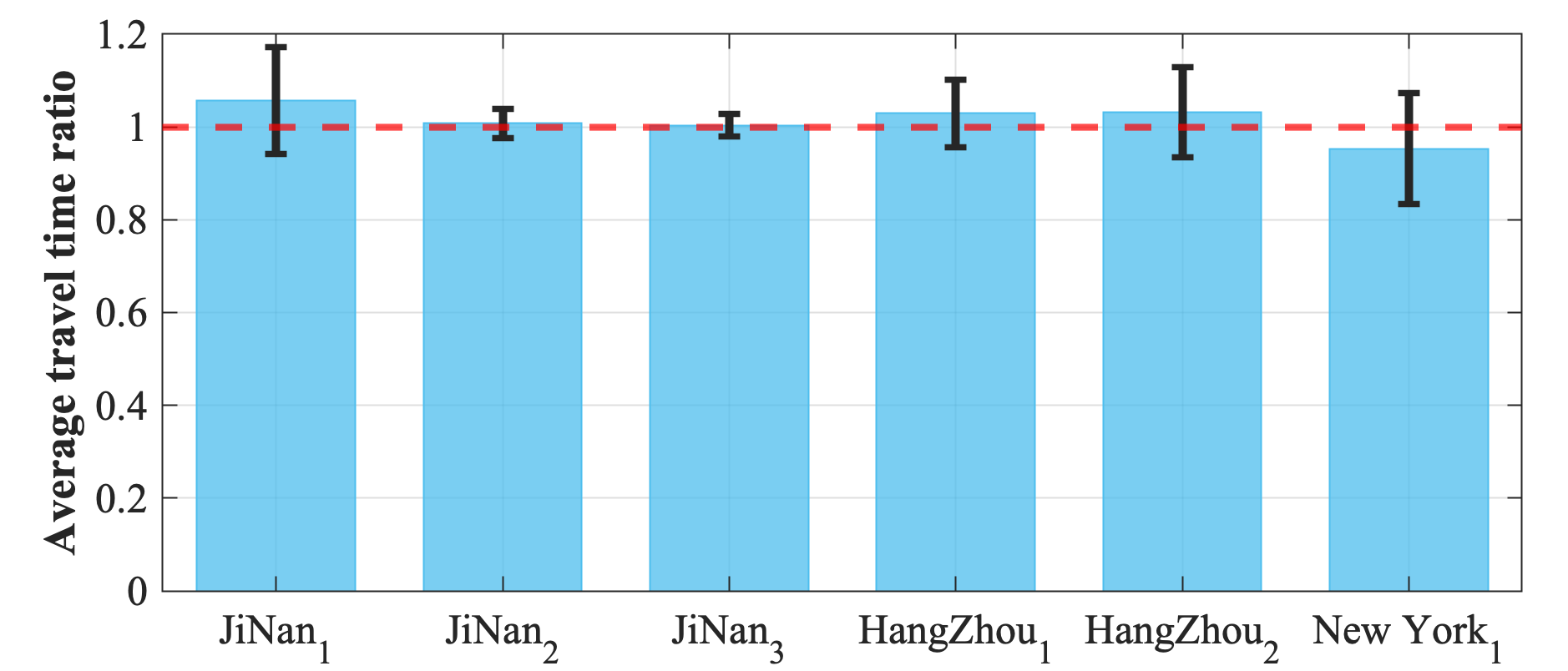}
     \caption{The average travel time of transfer divided by average travel time of direct training.  The error bars represent 95\% confident interval for average travel time ratio.}
     \label{fig:transfer}
\end{figure}

The transfer performance is denoted as an average travel time ratio: $t_{transfer}/t_{train}$, where $t_{transfer}$ and $t_{train}$ are the average travel time of transfer and direct training respectively.
The smaller of $t_{transfer}/t_{train}$, the better of the model transferability.
Figure~\ref{fig:transfer} reports the model transferability of Advanced-MPLight on all the datasets.
The closer the average travel time ratio is to one (red dashed line), the less degradation through the model transfer.
The average travel time ratio in New York is smaller than one, indicating that we can get better performance through transferring on it than direct training.
As is shown in Figure~\ref{fig:transfer}, Advanced-MPLight has high transfer performance over all the datasets, indicating that the model generalization of Advanced-MPLight is of great significance.
The results of model generalization indicate that Advanced-MPLight does not have an over-fitting problem in the RL training.

\section{Conclusion}
This paper proposes a novel method called Advanced-MP, based on MP and SOTL,
and designs an advanced traffic state (ATS) for the traffic movement representation with a pressure of queuing and demand of running for vehicles.
Experiments results on the large-scale road
networks with hundreds of traffic signals show that our ATS expresses more information for traffic state and boosts the performance of TSC methods.  

In the future, we will analyze more traffic factors and provide a more  precise traffic state representation to further optimize the TSC methods.

\section*{Acknowledgements}
This work is supported by the National Natural Science Foundation of China (Grant Nos. 61673150, 11622538). We acknowledge the Science Strength Promotion Programme of UESTC, Chengdu. 

\bibliography{references}
\bibliographystyle{icml2022}


\end{document}